\documentclass{article}

\usepackage[preprint]{neurips_2026}

\usepackage[utf8]{inputenc} 
\usepackage[T1]{fontenc}    
\usepackage{hyperref}       
\usepackage{url}            
\usepackage{booktabs}       
\usepackage{amsfonts}       
\usepackage{nicefrac}       
\usepackage{microtype}      
\usepackage{xcolor}         
\usepackage{graphicx}
\usepackage{amsmath,amssymb}
\usepackage{algorithm}
\usepackage{algpseudocode}
\usepackage{multirow}
\usepackage{subcaption}  
\usepackage[most]{tcolorbox}
\newtcolorbox{modelbox}[2][]{
  enhanced,
  breakable,
  colback=#2!6,
  colframe=#2!70!black,
  boxrule=0.8pt,
  arc=3pt,
  left=6pt,
  right=6pt,
  top=6pt,
  bottom=6pt,
  fonttitle=\bfseries,
  title=#1
}

\newtcolorbox{prompttemplate}{
  enhanced,
  breakable,
  colback=yellow!6,
  colframe=black,
  boxrule=0.8pt,
  arc=3pt,
  left=8pt,
  right=8pt,
  top=8pt,
  bottom=8pt,
  fontupper=\ttfamily\small,
}
\newcommand{\meanstd}[2]{#1{\scriptsize $\pm$#2}}
\title{Curriculum Learning-Guided Progressive Distillation in Large Language Models}

\author{
Jincheng Cao$^{^\dagger}$ \And
Fanzhi Zeng$^{^\dagger}$ \And
Leqi Liu$^{^\dagger}$ \And
Aryan Mokhtari$^{^\dagger,^\ddagger}$
}
\begin{document}

\maketitle

\begingroup
\renewcommand\thefootnote{}
\footnotetext{
\noindent
$^{^\dagger}$The University of Texas at Austin. 
$^{^\ddagger}$Google Research. \\
Corresponding author: Jincheng Cao <jinchengcao@utexas.edu>.
}
\endgroup

\begin{abstract}
  Knowledge distillation is a key technique for transferring the capabilities of large language models (LLMs) into smaller, more efficient student models. 
  Existing distillation approaches often overlook two critical factors: the learning order of training data and the capacity mismatch between teacher and student models. 
  This oversight limits distillation performance, as manifested by the counter-intuitive phenomenon where stronger teachers fail to produce better students. 
  In this work, we propose Curriculum Learning–Guided Progressive Distillation (CLPD), a unified framework that explicitly accounts for both factors by aligning data difficulty with teacher strength.
  CLPD constructs an explicit curriculum by organizing training examples from easy to hard, while simultaneously applying an implicit curriculum over supervision signals by progressively scheduling teachers of increasing capacity. 
  Our framework is modular and can be integrated into standard distillation algorithms with minimal overhead. 
  Empirical results on the reasoning benchmarks demonstrate that CLPD consistently outperforms standard distillation, data ordering alone, and teacher scheduling alone across multiple settings. 
  These findings highlight the importance of jointly considering data ordering and teacher capacity when distilling reasoning abilities into small language models.
\end{abstract}

\section{Introduction}

Knowledge distillation (KD) \cite{hinton2015distilling}, a widely used model compression paradigm in deep learning, addresses this challenge by transferring knowledge from a high-capacity teacher model to a smaller student model. As LLMs have advanced, KD has become a central strategy for building efficient small language models that preserve much of the teacher’s capability. This trend is evident in recent systems such as Llama 3.2 \cite{meta2024Llama}, Gemma-2 \cite{team2024gemma}, and the Deepseek-R1 distilled models \cite{deepseekai2025deepseekr1incentivizingreasoningcapability}, which explicitly rely on distillation to balance quality and efficiency.

In the LLM era, KD methods are often grouped into two broad families: white-box KD and black-box KD. White-box distillation refers to settings where the student has access to the teacher’s internal probability distributions or hidden states. Standard white-box methods align the teacher and student by matching their output distributions, often using Kullback–Leibler divergence (KLD) \cite{song2020lightpaff, gu2023minillm} or by aligning their intermediate representations \cite{sun2019patient, jiao2020tinybert, gupta2025efficient}. Recent studies \cite{gu2023minillm, agarwal2024policy} in white-box distillation highlight the importance of on-policy learning, where the student learns from its own generated responses rather than solely imitating the teacher’s outputs. These studies show that performing reverse KLD on student-generated text promotes mode-seeking behavior and reduces exposure bias compared to
teacher-forced training. However, white-box access is often impractical when the teacher is a proprietary API model (e.g., GPT-5). In this scenario, only teacher-generated texts are accessible, defining the more challenging black-box distillation setting. Black-box KD transfers knowledge through the teacher’s outputs \cite{kim2016sequence} and is widely adopted because many LLMs, such as GPT-4o \cite{hurst2024gpt}, Claude 3.5 \cite{claude35_sonnet_news}, and Gemini 1.5 \cite{team2024gemini}, are proprietary and expose only prediction interfaces. 

Despite these advances, most schemes pay limited attention to how training data and supervision are organized during distillation. In practice, poorly matched combinations of loss functions and data selection strategies can hinder student learning, especially when there is a substantial capacity gap between teacher and student models. Most KD methods treat teacher choice as fixed and training examples as i.i.d., rather than coupling teacher capability with example difficulty. Only a small number of works, such as \cite{xu2024speculative,liu2025being}, explicitly address aspects of this mismatch through modified supervision or curriculum-based training strategies during distillation.
Curriculum Learning (CL) exposes models to progressively more complex training examples and has been shown to improve optimization and generalization \cite{wu2020curricula,wang2021survey}. Recently, CL has been applied to LLM distillation \cite{liu2025being}; however, existing approaches typically focus on ordering training data while keeping the teacher fixed, leaving the interaction between data difficulty and teacher capability unexplored.

One curiously persistent phenomenon is that a better teacher does not always yield a stronger student. Prior works \cite{mirzadeh2020improved, jafari2021annealing, harutyunyan2023supervision, yim2026beyond} hypothesized that this arises from a capability gap \cite{zhang2025towards, kajitsuka2026revisiting} between teacher and student, and proposed progressive distillation, where the student is incrementally supervised by increasingly capable teachers. However, these approaches typically schedule teachers independently of the difficulty of training examples, leaving the interaction between teacher strength and data complexity largely unexplored.

\textbf{Contributions.}
In this paper, we argue that effective LLM distillation requires aligning what the student learns (data difficulty) with who teaches it (teacher capability). Inspired by human learning, where students first master fundamentals under simpler instruction before advancing to harder material, we propose Curriculum Learning–Guided Progressive Distillation (CLPD): an approach that gradually increases both training difficulty and teacher strength, reducing mismatch in the early stages while still providing high-quality guidance on the hardest examples.

Concretely, CLPD constructs a difficulty-ordered curriculum using CoT-derived complexity when available or student loss otherwise, and couples it with a teacher schedule that progresses from weaker to stronger models across curriculum stages. The method can be integrated seamlessly with the standard LLM distillation. Empirically, CLPD consistently outperforms strong distillation baselines as well as curriculum-only and progressive-only variants across different reasoning datasets. We complement these results with analyses that pinpoint when stronger teachers help and when they hurt, thereby clarifying the necessity of scheduling teachers in accordance with the data difficulty.

\section{Background and Preliminaries}

As our goal is to develop a plug-and-play framework for knowledge distillation, we first review common distillation objectives for LLMs, and then discuss two complementary factors that strongly affect distillation performance: the order in which training examples are presented and the capacity gap between teacher and student models.

\subsection{Knowledge Distillation}

Knowledge distillation was popularized by \cite{hinton2015distilling}, where the student model is trained to imitate the teacher’s token-level probability distribution. Let \( p_S^{\theta} \) denote the student distribution and \( p_T \) the teacher distribution. The standard supervised knowledge distillation loss \( \mathcal{L}_{\mathrm{SKD}} \) is defined as
%
\begin{align*}
    \mathcal{L}_{\mathrm{SKD}}(\theta)
:= \mathbb{E}_{x \sim X}
\Bigl[
\mathcal{D}_{\mathrm{KL}}\!\left( p_T(\cdot \mid x) \,\|\, p_S^{\theta}(\cdot \mid x) \right)
\Bigr]=
\mathbb{E}_{x \sim X}
\left[
\mathbb{E}_{y \sim p_T(\cdot \mid x)}
\left[
\log \frac{p_T(y \mid x)}{p_S^{\theta}(y \mid x)}
\right]
\right],
\end{align*}
where the outer expectation is taken over the input distribution and the inner expectation is taken over the teacher’s output distribution. This objective provides a dense training signal by leveraging the full token-level distribution of the teacher.
In contrast to this distribution-matching objective, \cite{kim2016sequence} introduced sequence-level knowledge distillation (SeqKD), which trains the student to maximize the likelihood of high-probability sequences generated by the teacher:
\begin{equation*}
\mathcal{L}_{\mathrm{SeqKD}}(\theta)
:= \mathbb{E}_{x \sim X}
\left[
\mathbb{E}_{y \sim p_T(\cdot \mid x)}
\left[
-\log p_S^{\theta}(y \mid x)
\right]
\right].
\end{equation*}
This can be viewed as supervised fine-tuning on teacher-generated outputs; in practice, the outputs $y$ are typically obtained via greedy or beam decoding rather than sampling. Sequence-level distillation is one of the most widely used distillation methods for large language models, as it does not require access to teacher logits and is computationally efficient. Standard single-teacher distillation with i.i.d.\ training data corresponds to the \emph{Vanilla} baseline shown in the last row of Fig.~\ref{fig:clpd}.

\subsection{Learning Order and Curriculum Learning}
A key challenge in KD for LLMs is the training--inference mismatch: student models are typically trained on fixed datasets that do not reflect the diverse distribution of inputs encountered at inference time \cite{gu2023minillm,agarwal2024policy}. To mitigate this issue, recent on-policy distillation methods incorporate student-generated outputs (SGOs) into training, allowing the student to learn from its own trajectories and reducing the gap between training and deployment. However, SGO usage introduces an inherent trade-off. While moderate SGO integration can improve generalization, excessive reliance on SGOs may degrade performance, as the distribution of noisy student-generated samples can deviate from the teacher’s training distribution and misguide the distillation process \cite{ko2024distillm, ko2025distillm}.

\begin{figure*}[t]
\vspace{-2mm}
    \centering
    \includegraphics[width=\linewidth]{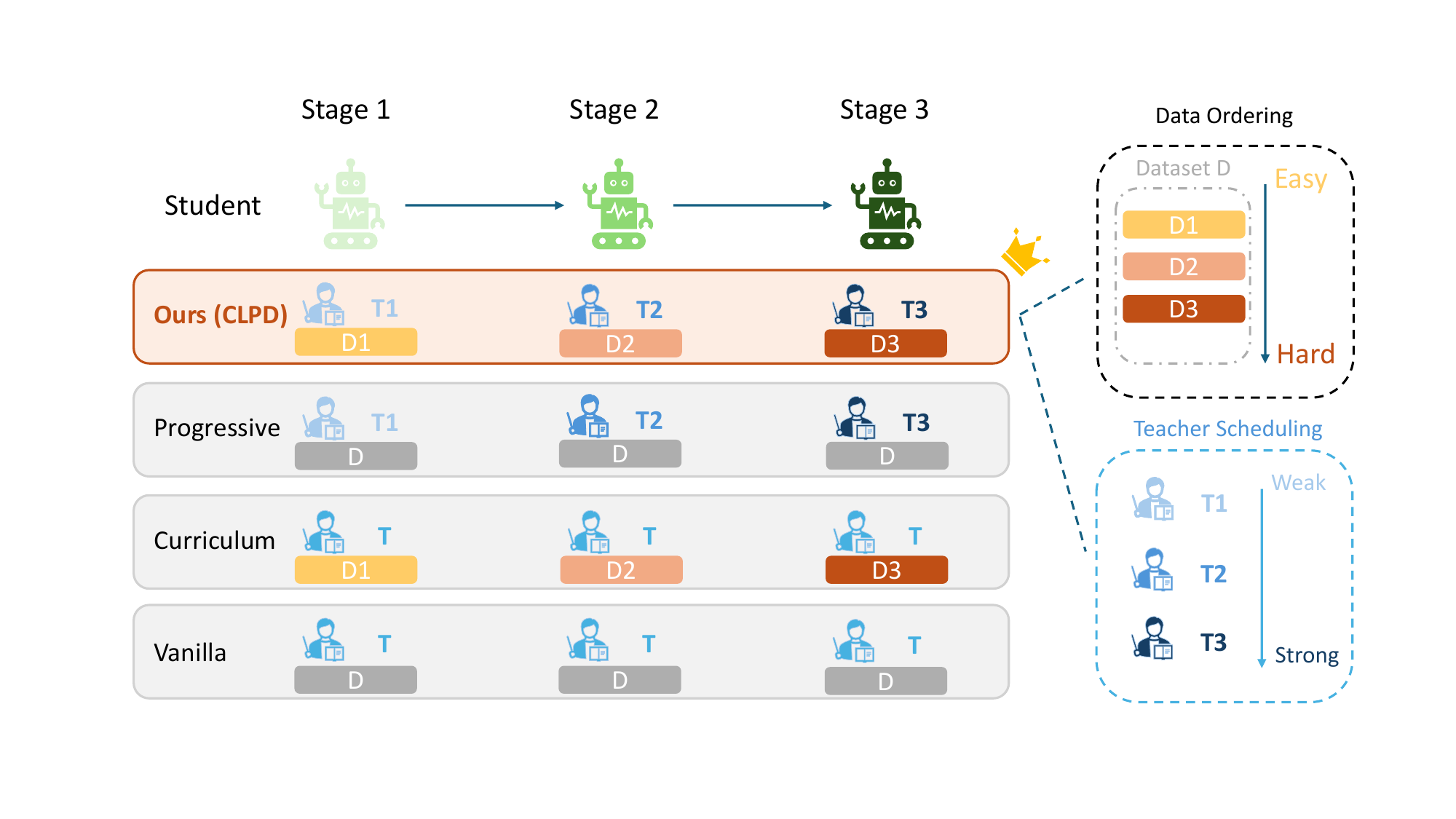}
    \vspace{-15mm}
    \caption{Overview of Curriculum Learning-Guided Progressive Distillation (CLPD) framework. CLPD combines an \textit{explicit curriculum} over data—organizing training examples from easy (D1) to hard (D3)—with an \textit{implicit curriculum} over supervision signals—progressively scheduling teachers from weak (T1) to strong (T3). 
    }
    \label{fig:clpd}
\end{figure*}

An alternative approach to addressing training--inference mismatch is to structure the distillation data in a meaningful learning order, drawing inspiration from curriculum learning. In the context of LLMs, curriculum learning \cite{wang2021survey} organizes training examples from simple to complex, mirroring human learning processes and enabling more stable optimization. A curriculum may begin with shorter prompts, simpler reasoning chains, or high-quality low-noise data, and gradually progress to longer, more syntactically complex, or noisier samples. Difficulty can be quantified using metrics such as prompt length, model loss, teacher confidence, or reasoning depth. Recent studies \cite{liu2025being, jiang2025teach} show that curriculum learning accelerates convergence, improves generalization, and enhances performance, particularly for instruction tuning and reasoning-intensive tasks, without increasing model size or dataset scale. However, curriculum learning mainly controls \emph{which examples} the student sees and \emph{when}, and does not address mismatches in the \emph{source} or \emph{difficulty} of supervision when the teacher is much stronger than the student. Applying curriculum learning with a fixed teacher corresponds to the \emph{Curriculum Learning only} baseline shown in the third row of Fig.~\ref{fig:clpd}.

\subsection{Capacity Gap and Progressive Distillation}

Given the capacity gap between teacher and student models, a stronger teacher does not necessarily yield a better student in knowledge distillation. When the teacher is too powerful, its supervision signal is often high-entropy or encodes nuanced multi-step reasoning that may exceed what a small student model can meaningfully absorb, leading to suboptimal dynamics and degraded student performance. Prior work \cite{mirzadeh2020improved, jafari2021annealing, harutyunyan2023supervision} has shown that, for a fixed student capacity, employing a weaker but more student-aligned teacher can be more effective, as such teachers induce distributions and task structures that are easier for the student to mimic.

An additional complication is that the teacher–student relationship is not static. As the student learns, its representations and predictions evolve, so a teacher that is initially well matched may become inadequate later in training. This motivates \emph{progressive distillation} \cite{panigrahi2024progressive, zhang2024minimal}, where supervision is staged by updating the teacher over time to maintain an appropriate difficulty level. The student thus learns first from simpler teachers to build core representations and later from stronger teachers as their capacity increases. A variety of strategies have been proposed for constructing intermediate teachers, including using intermediate checkpoints of a final teacher \cite{panigrahi2024progressive}, pruning or compressing a large teacher to obtain a spectrum of lighter-weight models \cite{zhang2024minimal}, or selecting pretrained models of increasing size from the same architecture family \cite{mirzadeh2020improved}. These approaches share the goal of keeping supervision commensurate with student capability over the course of training.

Scheduling multiple teachers over time while treating the training data stream as i.i.d.\ corresponds to the \emph{Progressive Distillation only} baseline shown in the second row of Fig.~\ref{fig:clpd}. However, existing progressive distillation frameworks typically schedule teachers as a function of training time or student progress, without explicitly accounting for the difficulty of individual training examples. As a result, they do not couple \emph{teacher capability} with \emph{example difficulty}: strong teachers may still be applied to easy instances early on, or weaker teachers to hard instances later in training, reintroducing supervision mismatch. This limitation motivates a unified approach that jointly schedules teachers and examples, aligning the strength of supervision with the difficulty of the data.

\section{The Proposed Framework}\label{sec:3}
We now present \emph{Curriculum Learning--Guided Progressive Distillation} (CLPD), a plug-and-play framework for LLM distillation on reasoning tasks. The key premise is that distillation is governed by \emph{two coupled axes}: (i) \emph{which examples} the student learns from and \emph{when} (learning order), and (ii) \emph{who teaches} those examples (teacher capability). Existing approaches typically control one axis while treating the other as fixed, which can leave a persistent supervision mismatch. CLPD resolves this by \emph{jointly} aligning \emph{example difficulty} with \emph{teacher strength} throughout training.

At a high level, CLPD proceeds in two stages. First, we construct an explicit curriculum by ranking training examples from easy to hard using a difficulty estimator. Second, we schedule multiple teachers along this curriculum so that weaker teachers supervise earlier, easier segments, and stronger teachers supervise later, harder segments. This coupling ensures that supervision remains within the student's effective learning range: the student is not overwhelmed by overly complex signals early on, while still receiving high-quality guidance on the hardest examples. The overall framework is presented in Fig.~\ref{fig:clpd}, and we describe each stage in detail below.


\subsection{Curriculum Construction}

CLPD begins by imposing structure on the training stream through an explicit easy-to-hard curriculum. The central objective is not simply to reorder data, but to control the \emph{pace} at which reasoning complexity is introduced. By exposing the student to simpler instances first, the model can acquire stable and reusable reasoning primitives before being asked to imitate longer or noisier chains of thought. This prepares the student to benefit from stronger supervision later in training.

We estimate example difficulty using one of two strategies, depending on the supervision signals available. When demonstration chain-of-thought (CoT) solutions are provided, we use their structural complexity as a direct and interpretable proxy for difficulty. Specifically, we rank examples primarily by the number of reasoning steps in the demonstrated solution: problems requiring fewer steps are treated as easier, while those requiring longer reasoning chains are treated as harder. When multiple examples share the same number of steps, we break ties using the total CoT length. This yields a transparent curriculum that closely reflects the depth of reasoning required by each problem.

When CoT demonstrations are not available, we instead estimate difficulty using the student model itself. In this setting, we run the (possibly warm-started) student on each training example and record the loss of its generated solution, treating lower-loss examples as easier and higher-loss examples as harder. Although this approach incurs an additional forward pass over the dataset, it provides a model-driven notion of difficulty that adapts to the student’s current representational capacity and does not rely on expert annotations.

After computing difficulty scores $\{d_i\}_{i=1}^N$ using either method, we sort the dataset into an ordered sequence $\mathcal{D}^{\mathrm{curr}}$ such that $d_{(1)} \le \cdots \le d_{(N)}$ and train on the data in this easy-to-hard order. In the student-loss setting, we compute difficulty scores using the initial (or warm-started) student and keep the ordering fixed during distillation for simplicity; while the curriculum could be refreshed dynamically, we find that a one-shot ordering is sufficient in practice. Importantly, this curriculum governs \emph{which examples} the student sees and \emph{when}, but it does not specify \emph{who} should supervise different parts of the curriculum. We address this missing dimension in the next subsection by introducing curriculum-aware teacher scheduling.

\begin{algorithm}[t!]
\caption{Curriculum Learning--Guided Progressive Distillation (CLPD)}
\label{alg:clpd}
\begin{algorithmic}[1]
\Require Dataset $\mathcal{D}$, teachers $\{T_j\}_{j=1}^n$, student $S_\theta$, threshold $\tau$, distillation algorithm $\mathcal{A}$

\State \textbf{Stage 1: Construct curriculum.}
\For{each $(x_i,y_i)\in\mathcal{D}$}
    \State Compute difficulty $d_i$ from CoT features if available; otherwise from student loss.
\EndFor
\State Sort $\mathcal{D}$ by increasing $d_i$ to obtain $\mathcal{D}^{\mathrm{curr}}$.

\State \textbf{Stage 2: Schedule teachers.}
\State Keep teachers $\mathcal{T}=\{T_j:\mathrm{Perf}(T_j)\ge\tau\}$ and order them weakest to strongest:
$\mathcal{T}=\{T^{(1)},\ldots,T^{(m)}\}$.
\State Split $\mathcal{D}^{\mathrm{curr}}$ into $m$ contiguous segments
$\mathcal{D}_1,\ldots,\mathcal{D}_m$ from easy to hard.

\For{$k=1$ to $m$}
    \For{each minibatch $B\subset\mathcal{D}_k$}
        \State $\theta \leftarrow \mathcal{A}(S_\theta,T^{(k)},B)$
    \EndFor
\EndFor

\State \Return trained student $S_\theta$
\end{algorithmic}
\end{algorithm}

\subsection{Teacher Scheduling}

Curriculum learning determines \emph{which examples} the student sees and \emph{when}, but not \emph{who} should supervise different parts of the curriculum. In LLM distillation, this missing dimension is often important: supervision from a very strong teacher can be overly complex early in training (e.g., high-entropy distributions or long reasoning chains), while supervision from a weak teacher can be insufficient on the hardest examples later on. As a result, using a single fixed teacher throughout training can create a persistent mismatch between supervision complexity and the student’s evolving competence, either overwhelming the student early or under-challenging it later.


CLPD addresses this issue by making teacher selection \emph{curriculum-aware}. The key idea is to align \emph{teacher capability} with \emph{example difficulty} so that supervision stays within the student’s effective learning range throughout training. Concretely, suppose we have access to $n$ candidate teacher models ${T_j}_{j=1}^n$ spanning different capability levels. We first evaluate each teacher on a small validation subset for the target task and retain only those whose performance exceeds a threshold $\tau$, yielding a viable set $\mathcal{T}$ of size $m$. This filtering step removes teachers that are too weak to provide reliable guidance.

Given the curriculum-ordered dataset $\mathcal{D}^{\mathrm{curr}}$, we then impose a monotone alignment between data difficulty and teacher strength. We order the viable teachers from weak to strong according to the performance metric $\mathrm{Perf}(\cdot)$, and partition $\mathcal{D}^{\mathrm{curr}}$ into $m$ (the number of viable teachers) contiguous segments of increasing difficulty. Each segment is assigned to one teacher, so that the student is first trained on easier examples under weaker supervision, and gradually transitions to stronger teachers as it encounters harder examples later in the curriculum. Contiguous segmentation preserves the easy-to-hard progression and prevents mixing easy and hard supervision within the same stage, which can otherwise introduce instability.

This staged teacher scheduling has two complementary effects. Early in training, weaker teachers provide supervision that the student can imitate more easily. 
Later in training, stronger teachers are introduced when the student is better prepared to learn more complex reasoning patterns, allowing the distillation process to exploit high-quality guidance on challenging examples. 
By explicitly coupling teacher strength with example difficulty, CLPD goes beyond curriculum-only and progressive-only strategies and resolves the supervision mismatch that arises when either axis is controlled in isolation. The complete procedure is described in Algorithm~\ref{alg:clpd}.

\subsection{Further Intuition and Empirical Interpretation}\label{sec:3.3}

We explain why CLPD is effective by linking its design to empirical behavior observed in our experiments. The central insight is that teacher optimality depends on \emph{example difficulty}: a teacher that is effective on hard instances can be suboptimal on easy ones, and vice versa. CLPD instantiates this principle by jointly scheduling training examples and teachers, enforcing a monotone alignment in which supervision strength increases only as the student progresses to more difficult data.

For a fixed student model, we attribute a teacher’s effectiveness in improving student performance via distillation on a given subset of examples to two factors: (i) \emph{task competence} on that subset, and (ii) \emph{teacher---student alignment}, i.e., how well the student can model the teacher’s output distribution for those examples. 
The ideal teacher model should strike a balance between the two. Strong teacher models typically have high competence, particularly on hard examples, but can be poorly aligned with the student and may provide complex supervision that a small student cannot reliably imitate early in training. 
On the other hand, weaker teachers may be better aligned with the student but lack sufficient competence, yielding low-quality supervision that caps student performance.



We validate the above explanation by distilling the same student model with different teachers on easy and hard subsets of GSM8K \cite{cobbe2021training}. Specifically, we use \texttt{Qwen2.5-1.5B-Instruct}, \texttt{Qwen2.5-3B-Instruct}, and \texttt{Qwen2.5-7B-Instruct} as teachers and \texttt{Qwen2.5-0.5B-Instruct} as the student. We perform standard distillation on either the top $20\%$ easiest or top $20\%$ hardest examples, and evaluate the resulting students on the full test set. To understand the underlying mechanism, we jointly measure (i) \emph{competence}, quantified by the teacher’s accuracy on each subset, (ii) \emph{alignment}, quantified by the teacher–student distribution gap via the student’s average negative log-probability on teacher-generated outputs, and (iii) the resulting student performance after distillation. The results are summarized in Table~\ref{tab:clpd_analysis}.

\begin{table*}[t!]
\centering
\caption{Teacher effectiveness on GSM8K across difficulty splits. We report (i) competence (accuracy), (ii) teacher--student alignment (student loss), and (iii) student performance (Qwen-0.5B) after distillation only on easy or hard subsets. Bold values highlight favorable results for each metric.}
\label{tab:clpd_analysis}

\footnotesize
\setlength{\tabcolsep}{6pt}
\renewcommand{\arraystretch}{1.1}

\begin{tabular}{lcccccc}
\toprule
\multirow{2}{*}{\textbf{Teacher}} 
& \multicolumn{2}{c}{\textbf{Teacher Accuracy $\uparrow$}} 
& \multicolumn{2}{c}{\textbf{Teacher-student Alignment $\downarrow$}} 
& \multicolumn{2}{c}{\textbf{Student Accuracy $\uparrow$}} \\
\cmidrule(lr){2-3} \cmidrule(lr){4-5} \cmidrule(lr){6-7}
& \textbf{Hard} & \textbf{Easy} 
& \textbf{Hard} & \textbf{Easy} 
& \textbf{Hard} & \textbf{Easy} \\
\midrule
Qwen-1.5B 
& 0.611 & 0.855 
& 0.373 & \textbf{0.229} 
& 0.394 & 0.418 \\

Qwen-3B 
& 0.823 & \textbf{0.928} 
& 0.352 & \textbf{0.230} 
& 0.420 & \textbf{0.440} \\

Qwen-7B 
& \textbf{0.906} & \textbf{0.967} 
& \textbf{0.340} & 0.235 
& \textbf{0.424} & 0.432 \\
\bottomrule
\end{tabular}
\end{table*}

Empirically, we observe that the strongest teacher (7B) yields the best student on the hard subset, while a mid-sized teacher (3B) yields the best student on the easy subset.
On the hard subset, the 7B teacher is more competent and more aligned to the student relative to other teachers. 
On the easy subset, the 7B and 3B teachers have comparable competence, but the 7B teacher is less aligned with the student, resulting in the 3B teacher being more effective.
Importantly, high teacher–student alignment alone does not make a teacher model better: although the smallest teacher (1.5B) is more aligned with the student on easy data, its lower competence still leads to inferior student outcomes.

Overall, these results support the central design principle of CLPD: effective distillation requires selecting teachers as a function of example difficulty by balancing competence and alignment. This directly motivates curriculum-aware teacher scheduling, which avoids both early-stage overload from overly strong supervision and late-stage under-supervision on the hardest examples.

\section{Experiments}\label{sec:experiments}

We evaluate CLPD on mathematical reasoning and commonsense reasoning tasks, focusing on two questions: 
(i) whether aligning teacher capability with example difficulty improves over standard single-teacher distillation, and 
(ii) whether the gains come from the \emph{coupling} of curriculum learning and progressive distillation rather than either component alone. 
We evaluate both within- and cross-family distillation under a fixed training budget, comparing CLPD with strong baselines under identical optimization and data conditions.

Our primary distillation framework is sequence-level knowledge distillation (SeqKD)~\cite{kim2016sequence}, where the student is trained on teacher-generated chain-of-thought (CoT) demonstrations. 
Within this framework, we compare CLPD against three baselines: 
(i) standard SeqKD with a single fixed teacher, 
(ii) curriculum learning only, where training examples are ordered from easy to hard while the teacher remains fixed, and 
(iii) progressive distillation only, where multiple teachers are scheduled over training while the data are treated as i.i.d. 
These baselines allow us to isolate the contributions of data ordering, teacher scheduling, and their interaction.

\begin{table*}[t!]
\centering
\caption{Performance (Mean $\pm$ Std, n=3) for different teacher--student configurations on the GSM8K and MATH datasets.
Bold rows denote our CLPD method.}
\label{tab:combined_results}

\footnotesize
\setlength{\tabcolsep}{6pt}
\renewcommand{\arraystretch}{1.05}
\setlength{\aboverulesep}{0pt}
\setlength{\belowrulesep}{0pt}

\begin{tabular}{lcccc}
\toprule
\multirow{2}{*}{\textbf{Configuration}}
 & \multicolumn{2}{c}{\textbf{GSM8K}} 
 & \multicolumn{2}{c}{\textbf{MATH}} \\
\cmidrule(lr){2-3} \cmidrule(lr){4-5}
 & \textbf{Qwen-0.5B} & \textbf{Llama-1B}
 & \textbf{Qwen-3B} & \textbf{Llama-3B} \\
\midrule

\multicolumn{5}{l}{\emph{Initial models}} \\
student
& 0.414 & 0.406 & 0.475 & 0.038 \\
small teacher 
& \multicolumn{2}{c}{0.885} & \multicolumn{2}{c}{0.646} \\
large teacher 
& \multicolumn{2}{c}{0.949} & \multicolumn{2}{c}{0.690} \\

\midrule
\multicolumn{5}{l}{\emph{Standard distillation}} \\
$\leftarrow$ small teacher 
& \meanstd{0.449}{0.008} & \meanstd{0.427}{0.010} & \meanstd{0.550}{0.006} & \meanstd{0.130}{0.011} \\
$\leftarrow$ large teacher 
& \meanstd{0.446}{0.009} & \meanstd{0.424}{0.006} & \meanstd{0.546}{0.004} & \meanstd{0.128}{0.013} \\

\midrule
\multicolumn{5}{l}{\emph{Curriculum learning only}} \\
$\leftarrow$ small teacher + order (expert CoT) 
& \meanstd{0.457}{0.007} & \meanstd{0.446}{0.007} & \meanstd{0.550}{0.009} & \meanstd{0.130}{0.007} \\
$\leftarrow$ small teacher + order (student-loss) 
& \meanstd{0.452}{0.006} & \meanstd{0.438}{0.005} & \meanstd{0.552}{0.008} & \meanstd{0.134}{0.006} \\
$\leftarrow$ large teacher + order (expert CoT) 
& \meanstd{0.451}{0.009} & \meanstd{0.440}{0.007} & \meanstd{0.554}{0.006} & \meanstd{0.129}{0.010} \\
$\leftarrow$ large teacher + order (student-loss) 
& \meanstd{0.451}{0.006} & \meanstd{0.444}{0.009} & \meanstd{0.553}{0.005} & \meanstd{0.134}{0.006} \\

\midrule
\multicolumn{5}{l}{\emph{Progressive distillation only}} \\
$\leftarrow$ small $\&$ large teacher 
& \meanstd{0.460}{0.006} & \meanstd{0.449}{0.008} & \meanstd{0.550}{0.007} & \meanstd{0.139}{0.007} \\

\midrule
\multicolumn{5}{l}{\textbf{CLPD (ours)}} \\
\textbf{$\leftarrow$ order (expert CoT)} 
& \textbf{\meanstd{0.485}{0.010}} & \textbf{\meanstd{0.465}{0.008}} & \textbf{\meanstd{0.564}{0.008}} & \textbf{\meanstd{0.150}{0.007}} \\
\textbf{$\leftarrow$ order (student-loss)} 
& \textbf{\meanstd{0.482}{0.007}} & \textbf{\meanstd{0.463}{0.009}} & \textbf{\meanstd{0.566}{0.005}}  & \textbf{\meanstd{0.153}{0.011}} \\

\bottomrule
\end{tabular}
\end{table*}
\subsection{Mathematical Reasoning}


\textbf{Setup.} We first evaluate CLPD on two standard math reasoning benchmarks. GSM8K~\cite{cobbe2021training} features grade-school arithmetic problems with CoT solutions, while MATH~\cite{hendrycks2024measuring} includes more challenging high-school and competition-level problems across diverse domains. Together, they span a wide range of reasoning difficulty and test CLPD under varying task complexity. We adopt the two data ordering strategies in Algorithm~\ref{alg:clpd} to rank the training data and partition it evenly into easy and hard subsets. Our experiments focus on improving the reasoning performance of relatively small language models, where distillation is most impactful. We consider student models from two model families: Qwen-2.5~\cite{yang2025qwen3} and Llama-3.2~\cite{grattafiori2024Llama}. For GSM8K, we use \texttt{Qwen2.5-0.5B-Instruct} and \texttt{Llama3.2-1B-Instruct} as students; for MATH, we use \texttt{Qwen2.5-3B} and \texttt{Llama3.2-3B}. As teacher models, we adopt \texttt{Qwen2.5-3B-Instruct} and \texttt{Qwen2.5-7B-Instruct} for GSM8K, and \texttt{Qwen2.5-7B-Instruct} and \texttt{Qwen2.5-14B-Instruct} for MATH. We restrict our experiments to two teachers per task, as this setting already captures the key effects of progressive supervision and yields consistent improvements. In general, larger Llama models show weaker math reasoning than their Qwen counterparts. We therefore use Qwen models as teachers, while including both Qwen and Llama models as students to evaluate CLPD under both within-family and cross-family distillation.




\textbf{Results.} Table~\ref{tab:combined_results} summarizes the performance of CLPD against standard distillation, curriculum learning, and progressive distillation. Across both Qwen and Llama setups on GSM8K and MATH, student models start far below their teachers, underscoring the difficulty of transferring reasoning capabilities. \emph{Standard distillation} yields only limited gains, and stronger teachers do not consistently improve outcomes. \emph{Curriculum learning} (easy-to-hard ordering) and \emph{progressive distillation} (multi-teacher scheduling) each provide modest improvements, but remain insufficient when used alone. In contrast, \emph{CLPD (ours)} consistently achieves the best performance across all settings. Compared to the strongest standard distillation baseline, CLPD improves Qwen student from 0.449 to 0.485 on GSM8K and from 0.550 to 0.566 on MATH, with similar gains observed for Llama students. By aligning data difficulty with teacher capability, it outperforms all baselines and remains robust across different curriculum construction strategies. These results provide evidence that neither data ordering nor teacher scheduling alone is sufficient; the strongest gains arise when the two are aligned.

\subsection{Commonsense Reasoning}

\textbf{Setup.} We extend CLPD to commonsense reasoning tasks, where models must perform multi-hop inference and leverage implicit world knowledge rather than explicit computation. As in the mathematical setting, all experiments follow a sequence-level knowledge distillation (SeqKD) framework~\cite{kim2016sequence}, training student models on teacher-generated chain-of-thought (CoT) demonstrations. We evaluate on StrategyQA~\cite{geva2021did} and ARC-Challenge~\cite{clark2018think}. StrategyQA requires implicit multi-step reasoning over commonsense knowledge, while ARC-Challenge consists of difficult multiple-choice science questions that challenge shallow reasoning methods. Unlike math datasets, these benchmarks lack expert CoT, so we estimate example difficulty using student loss as in Algorithm~\ref{alg:clpd}. We focus on improving small language models, using \texttt{Qwen2.5-0.5B} and \texttt{Llama3.2-1B} as students. For teachers, we use \texttt{Qwen2.5-3B-Instruct}/\texttt{Qwen2.5-7B-Instruct} on StrategyQA and \texttt{Qwen2.5-7B-Instruct}/\texttt{Qwen2.5-14B-Instruct} on ARC-Challenge.


\begin{table*}[t!]
\centering
\caption{Performance (Mean $\pm$ Std, n=3) for different teacher--student configurations on the StrategyQA and ARC-Challenge datasets.
Bold rows denote our CLPD method.}
\label{tab:commonsense_results}

\footnotesize
\setlength{\tabcolsep}{6pt}
\renewcommand{\arraystretch}{1.05}
\setlength{\aboverulesep}{0pt}
\setlength{\belowrulesep}{0pt}

\begin{tabular}{lcccc}
\toprule
\multirow{2}{*}{\textbf{Configuration}}
 & \multicolumn{2}{c}{\textbf{StrategyQA}} 
 & \multicolumn{2}{c}{\textbf{ARC-Challenge}} \\
\cmidrule(lr){2-3} \cmidrule(lr){4-5}
 & \textbf{Qwen-0.5B} & \textbf{Llama-1B}
 & \textbf{Qwen-0.5B} & \textbf{Llama-1B} \\
\midrule

\multicolumn{5}{l}{\emph{Initial models}} \\
student
& 0.253 & 0.039 & 0.171 & 0.077 \\
small teacher 
& \multicolumn{2}{c}{0.659} & \multicolumn{2}{c}{0.926} \\
large teacher 
& \multicolumn{2}{c}{0.696} & \multicolumn{2}{c}{0.943} \\

\midrule
\multicolumn{5}{l}{\emph{Standard distillation}} \\
$\leftarrow$ small teacher 
& \meanstd{0.518}{0.011} & \meanstd{0.541}{0.012} & \meanstd{0.454}{0.005} & \meanstd{0.347}{0.014} \\
$\leftarrow$ large teacher 
& \meanstd{0.514}{0.012} & \meanstd{0.542}{0.013} & \meanstd{0.455}{0.006} & \meanstd{0.345}{0.014} \\

\midrule
\multicolumn{5}{l}{\emph{Curriculum learning only}} \\
$\leftarrow$ small teacher + order (student-loss) 
& \meanstd{0.521}{0.008} & \meanstd{0.551}{0.010} & \meanstd{0.469}{0.010} & \meanstd{0.355}{0.012} \\
$\leftarrow$ large teacher + order (student-loss) 
& \meanstd{0.524}{0.009} & \meanstd{0.544}{0.011} & \meanstd{0.465}{0.014} & \meanstd{0.356}{0.015} \\

\midrule
\multicolumn{5}{l}{\emph{Progressive distillation only}} \\
$\leftarrow$ small $\&$ large teacher
& \meanstd{0.526}{0.007} & \meanstd{0.543}{0.005} & \meanstd{0.463}{0.005} & \meanstd{0.362}{0.005} \\

\midrule
\multicolumn{5}{l}{\textbf{CLPD (ours)}} \\
\textbf{$\leftarrow$ order (student-loss)} 
& \textbf{\meanstd{0.555}{0.013}} & \textbf{\meanstd{0.563}{0.014}} & \textbf{\meanstd{0.488}{0.009}} & \textbf{\meanstd{0.374}{0.013}} \\

\bottomrule
\end{tabular}
\end{table*}

\textbf{Results.} The results are presented in Table~\ref{tab:commonsense_results}. Across both datasets, student models start far below their teachers, highlighting the difficulty of transferring reasoning capabilities. \emph{Standard distillation} provides moderate gains, but using a stronger teacher does not consistently improve performance. \emph{Curriculum learning} and \emph{progressive distillation} each yield incremental improvements over standard distillation, but their gains remain limited when applied independently. In contrast, \emph{CLPD (ours)} consistently achieves the best performance across all settings. Compared to the best standard distillation baseline, CLPD improves Qwen student from 0.518 to 0.555 on StrategyQA and from 0.455 to 0.488 on ARC-Challenge, with similar gains for Llama students. By aligning data difficulty with teacher capability, CLPD effectively reduces supervision mismatch and leads to stronger and more consistent improvements. The low initial performance on commonsense tasks is largely caused by strict final-answer formatting requirements: base models often produce reasoning but fail to end with the required answer format. Distillation improves both reasoning and instruction following; we provide details in Appendix~\ref{app:setup}. Having observed consistent gains across both mathematical and commonsense reasoning, we next examine whether CLPD is sensitive to how the curriculum is partitioned across teachers.

\subsection{Robustness to Curriculum Partitioning}\label{sec:partition}

In our default implementation, data are uniformly partitioned across teachers for simplicity and stability, yielding a monotonic easy-to-hard curriculum. We now examine whether CLPD depends on this specific partitioning choice by varying the curriculum split at which supervision transitions from the weaker teacher to the stronger teacher. For consistency, we use Qwen models and rank data difficulty using student loss.

\begin{figure}[t]
    \centering
    \begin{subfigure}[t]{0.24\linewidth}
        \centering
        \includegraphics[width=\linewidth]{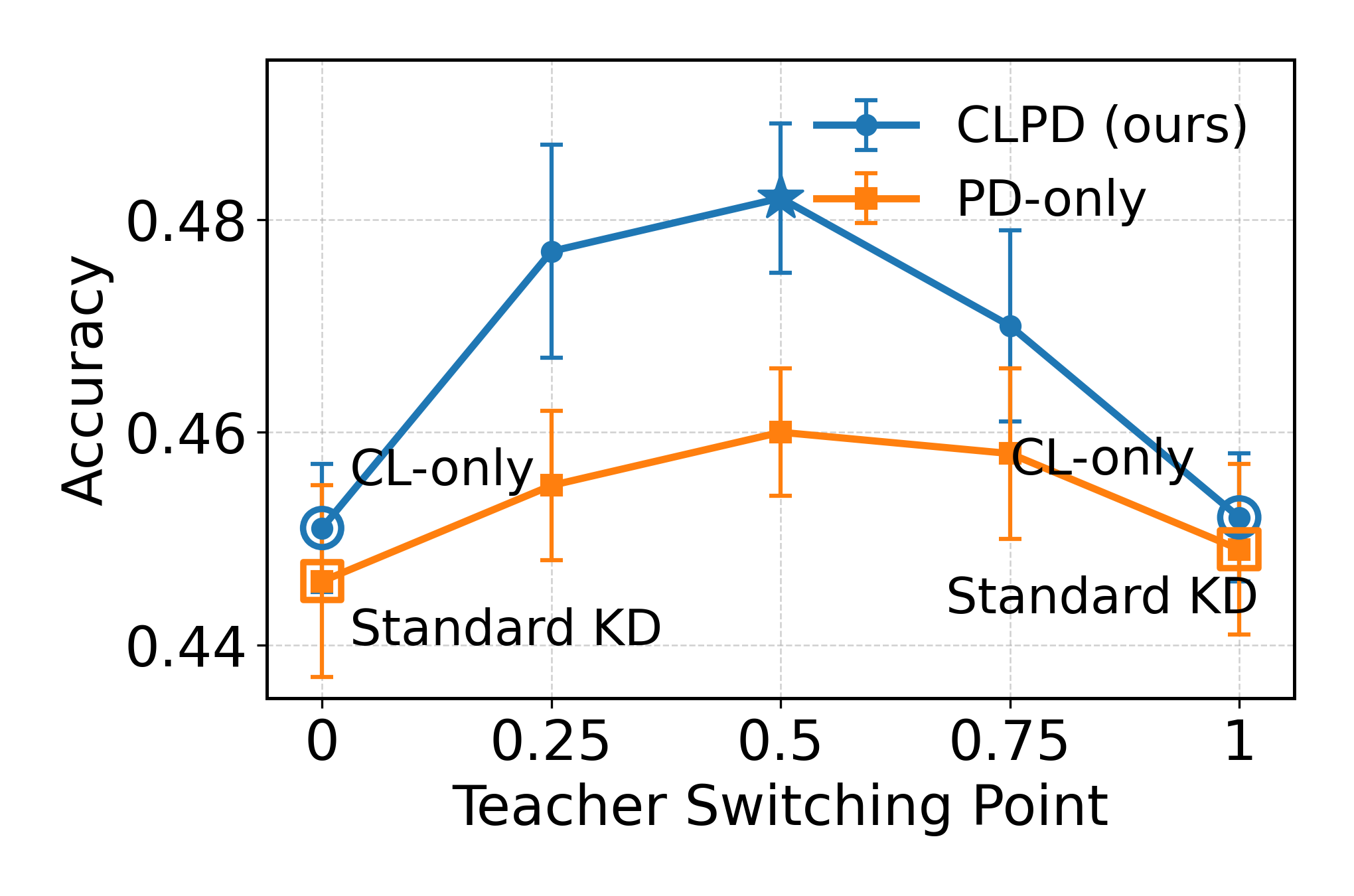}
        \caption{GSM8K}
    \end{subfigure}
    \hfill
    \begin{subfigure}[t]{0.24\linewidth}
        \centering
        \includegraphics[width=\linewidth]{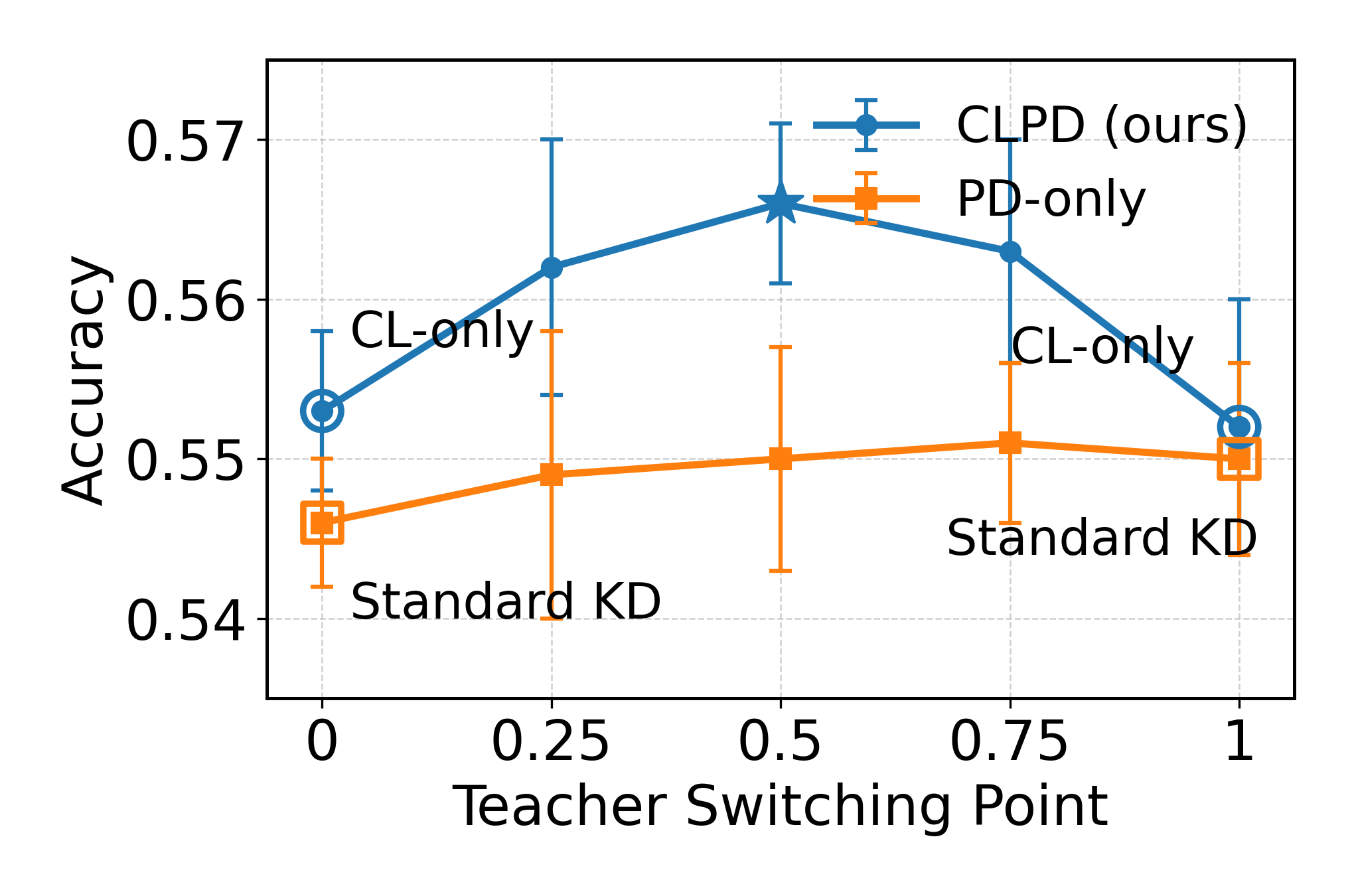}
        \caption{MATH}

    \end{subfigure}
    \hfill
    \begin{subfigure}[t]{0.24\linewidth}
        \centering
        \includegraphics[width=\linewidth]{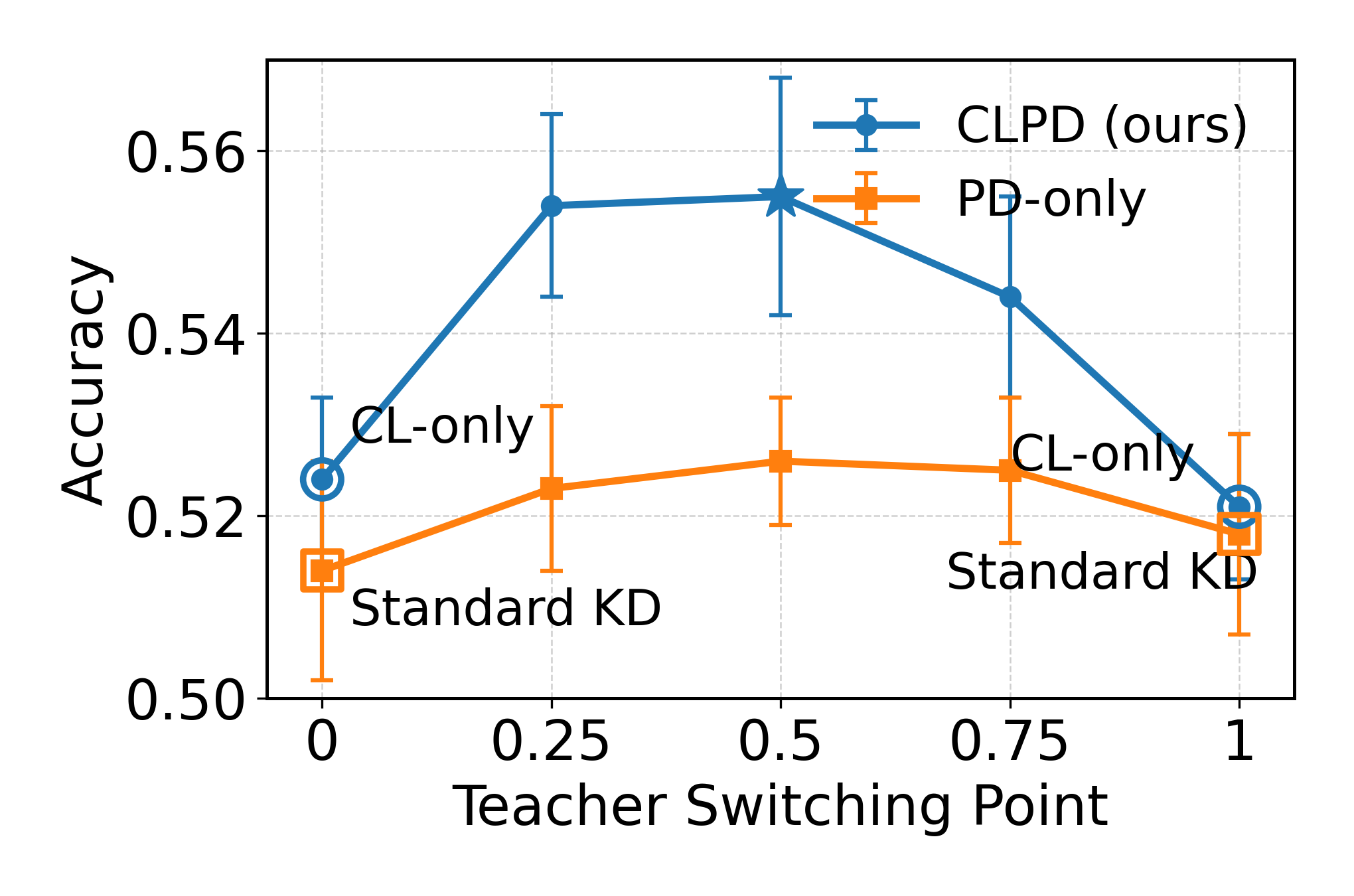}
        \caption{StrategyQA}

    \end{subfigure}
    \hfill
    \begin{subfigure}[t]{0.24\linewidth}
        \centering
        \includegraphics[width=\linewidth]{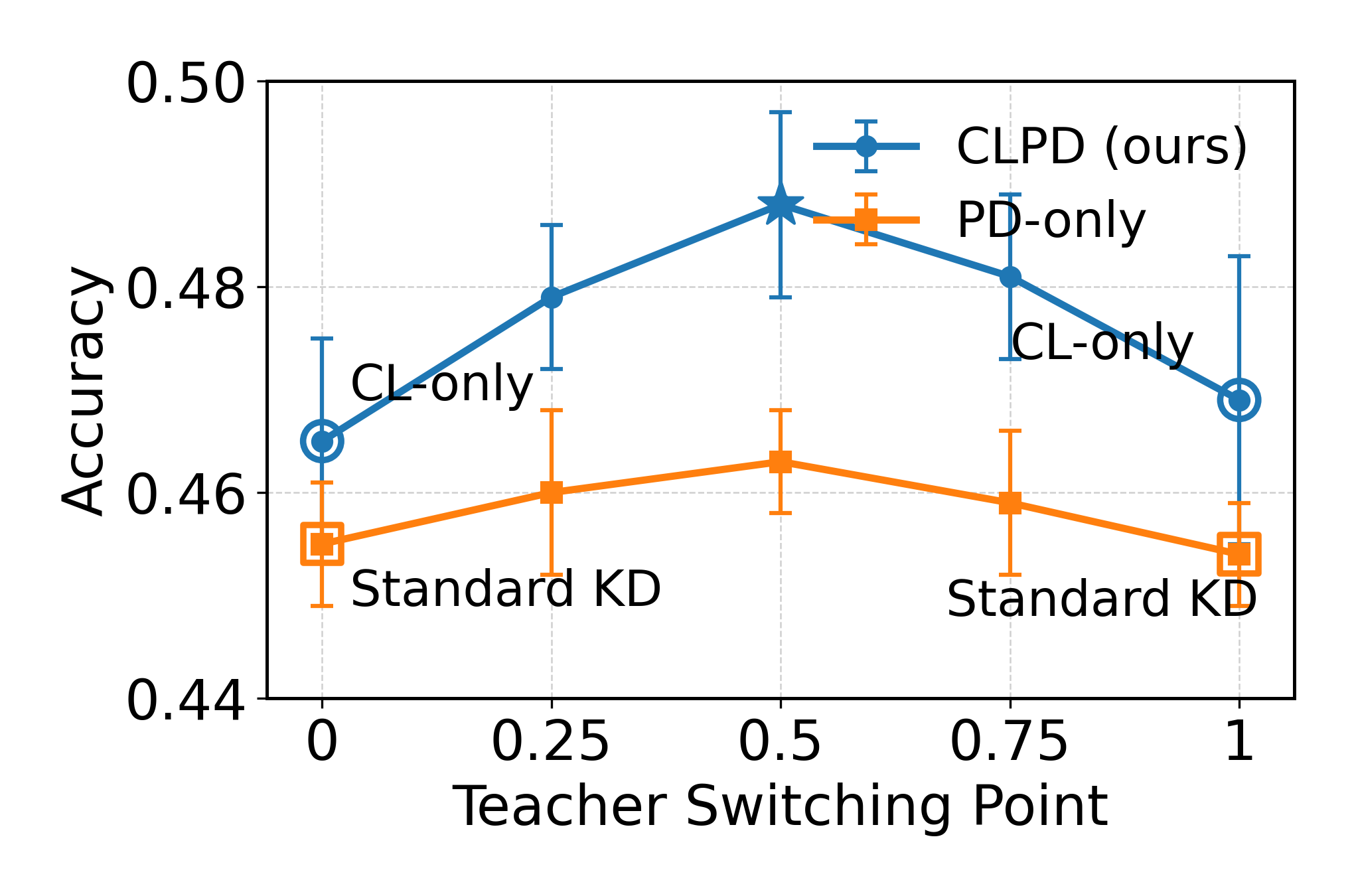}
        \caption{ARC-Challenge}

    \end{subfigure}
    
    \caption{CLPD vs.\ PD under different data partitions.}
    \label{fig:clpd_partition}
\end{figure}

Figure~\ref{fig:clpd_partition} shows that CLPD consistently outperforms progressive distillation (PD) across different partitions on GSM8K, MATH, StrategyQA, and ARC-Challenge. Notably, CLPD remains robust under non-uniform partitions, with performance varying only mildly across a wide range of splits. The best results are often achieved near balanced splits, but the overall trend indicates that CLPD does not rely on precise tuning of the partition.

In contrast, PD exhibits weaker and less stable performance, suggesting that teacher scheduling alone is insufficient without accounting for data difficulty. These results highlight that the primary driver of improvement is not the exact switching point, but the alignment between teacher capability and example difficulty. As long as weaker teachers are applied to easier examples and stronger teachers to harder ones, CLPD consistently yields strong performance.

\subsection{Ablation Study}

The robustness results above suggest that precise partitioning is not the key factor; rather, performance depends on aligning data difficulty with teacher capability. To isolate this effect, we conduct a coupling ablation that breaks this alignment while keeping the same teachers, training budget, and partitioning. We consider two variants. In CLPD-RT, we keep the easy-to-hard data order but reverse the teacher schedule (large $\rightarrow$ small), assigning stronger teachers to easier examples and weaker teachers to harder ones. In CLPD-RD, we keep the teacher schedule (small $\rightarrow$ large) but reverse the data order (hard $\rightarrow$ easy), exposing harder examples earlier in training. For consistency, we use Qwen models and rank data difficulty using student loss.

As shown in Table~\ref{tab:coupling_ablation}, both variants underperform CLPD. Although each retains one component (curriculum or progressive distillation), breaking their alignment degrades performance. These results confirm that CLPD’s gains come not simply from combining curriculum learning and progressive distillation, but from \emph{explicitly coupling} them through difficulty-aware teacher scheduling. 
Effective distillation requires data difficulty and teacher capacity to progress in tandem.

\begin{table*}[t!]
\centering
\caption{Coupling ablation of CLPD by reversing teacher schedule or data order.}

\label{tab:coupling_ablation}
\footnotesize
\setlength{\tabcolsep}{6pt}
\renewcommand{\arraystretch}{1.05}
\begin{tabular}{lcccccc}
\toprule
\textbf{Method} & \textbf{Data Order} & \textbf{Teacher Schedule} & \textbf{GSM8K} & \textbf{MATH} & \textbf{StrategyQA} & \textbf{ARC-Challenge} \\
\midrule
CLPD
& easy $\rightarrow$ hard 
& small $\rightarrow$ large 
& \textbf{\meanstd{0.482}{0.007}} & \textbf{\meanstd{0.566}{0.005}} & \textbf{\meanstd{0.555}{0.013}} & \textbf{\meanstd{0.488}{0.009}} \\
\midrule
CLPD-RT 
& easy $\rightarrow$ hard 
& large $\rightarrow$ small 
& \meanstd{0.451}{0.011} & \meanstd{0.552}{0.007} & \meanstd{0.525}{0.015} & \meanstd{0.463}{0.008} \\
CLPD-RD     
& hard $\rightarrow$ easy 
& small $\rightarrow$ large 
& \meanstd{0.455}{0.009} & \meanstd{0.549}{0.009} & \meanstd{0.521}{0.007} & \meanstd{0.461}{0.011}\\
\bottomrule
\end{tabular}
\end{table*}

\section{Conclusion}\label{sec:conclusion}

We introduced Curriculum Learning–Guided Progressive Distillation (CLPD), a unified and plug-and-play framework that improves knowledge distillation by aligning data difficulty with teacher supervision strength. By jointly ordering training examples and progressively scheduling teachers, CLPD reduces supervision mismatch and enables more effective knowledge transfer without architectural changes or additional computational overhead. Across multiple reasoning benchmarks, CLPD consistently outperforms standard distillation, curriculum learning alone, and progressive distillation alone. More broadly, our findings suggest that many distillation failures attributed to teacher quality instead stem from misalignment between supervision strength and learning stage, and that explicitly coupling these factors is essential for transferring reasoning capabilities to smaller language models. 
\bibliographystyle{unsrt}
\bibliography{ref.bib}






\appendix

\newpage

\section{Beyond Two Teachers}\label{app:three_teachers}

We further study CLPD in settings with more than two teachers. Specifically, we consider a three-teacher setup with increasing model capacity, where the student is trained progressively from smaller to larger teachers. As shown in Table~\ref{tab:3_teachers}, using three teachers can provide improvements over the two-teacher setting in most cases.

Overall, the results suggest that CLPD can benefit from incorporating additional teachers, as this allows for a more fine-grained alignment between data difficulty and teacher capability throughout training. In practice, however, two well-chosen teachers are often sufficient to capture the primary benefits of this alignment, offering a favorable balance between performance and simplicity while avoiding the added complexity of more intricate scheduling strategies.



\begin{table*}[t]
\centering
\caption{Performance of CLPD with three teachers across different student models and datasets. Numbers in parentheses denote the change relative to the two-teacher CLPD.}
\label{tab:3_teachers}

\footnotesize 
\setlength{\tabcolsep}{6pt}
\renewcommand{\arraystretch}{1.05}

\begin{tabular}{lcccc}
\toprule
\textbf{Student Model} 
& \textbf{GSM8K} 
& \textbf{MATH}
& \textbf{StrategyQA}
& \textbf{ARC-Challenge} \\
\midrule
\textbf{Qwen} 
& 0.489 {\scriptsize (+0.007)}
& 0.565 {\scriptsize (-0.001)}
& 0.559 {\scriptsize (+0.004)}
& 0.498 {\scriptsize (+0.010)} \\

\textbf{Llama} 
& 0.470 {\scriptsize (+0.007)}
& 0.157 {\scriptsize (+0.004)}
& 0.560 {\scriptsize (-0.003)}
& 0.381 {\scriptsize (+0.007)} \\
\bottomrule
\end{tabular}
\end{table*}

\section{Static vs.\ Dynamic Curriculum Construction}
\label{app:static_curriculum}

In CLPD, we construct the curriculum using a static (one-shot) difficulty ordering computed prior to training and kept fixed throughout distillation. A natural question is whether incorporating adaptivity—by updating difficulty estimates as the student evolves—can further improve performance.

To explore this, we consider a simple two-stage re-ranking scheme. We first compute a global difficulty ordering using the loss of a pretrained (or warm-started) student model. During training, after the first stage is completed, we re-estimate difficulty using the updated student and re-rank the remaining portion of the data (i.e., the second half). This provides a lightweight approximation to a dynamic curriculum while preserving most of the original training pipeline.

\begin{table*}[t!]
\centering
\caption{Performance of CLPD under dynamic curriculum (two-stage re-ranking). Numbers in parentheses denote changes relative to the static curriculum (one-shot).}
\label{tab:order_twice}

\footnotesize 
\setlength{\tabcolsep}{6pt}
\renewcommand{\arraystretch}{1.05}

\begin{tabular}{lcccc}
\toprule
\textbf{Student Model} 
& \textbf{GSM8K} 
& \textbf{MATH}
& \textbf{StrategyQA}
& \textbf{ARC-Challenge} \\
\midrule
\textbf{Qwen} 
& 0.485 {\scriptsize (+0.003)}
& 0.561 {\scriptsize (-0.005)}
& 0.554 {\scriptsize (-0.001)}
& 0.491 {\scriptsize (+0.003)} \\

\textbf{Llama} 
& 0.465 {\scriptsize (+0.002)}
& 0.149 {\scriptsize (-0.004)}
& 0.568 {\scriptsize (+0.005)}
& 0.371 {\scriptsize (-0.003)} \\
\bottomrule
\end{tabular}
\end{table*}




As shown in Table~\ref{tab:order_twice}, the two-stage re-ranking scheme yields mixed results, with small gains on some datasets but no consistent improvement over the static curriculum. This suggests that a one-shot ordering already captures most of the useful difficulty structure for CLPD.

One reason is that the relative difficulty of examples is largely stable: instances that are difficult for the initial student tend to remain challenging throughout training. Moreover, CLPD primarily relies on coarse alignment between data difficulty and teacher capability, making it robust to moderate inaccuracies in the exact ordering.

While dynamic curricula can in principle better track the student’s evolving capabilities, they introduce additional computational overhead due to repeated evaluation and re-ranking. They may also disrupt the monotonic easy-to-hard progression, potentially interfering with the intended teacher scheduling.

Overall, these results suggest that CLPD does not rely on precise or frequently updated difficulty estimates. A simple static curriculum provides a strong and stable signal, capturing most of the benefits while maintaining an efficient and straightforward training procedure. Such one-shot difficulty estimation is also commonly adopted in prior work~\cite{liu2025being, wang2025angles}.

\section{CLPD and On-policy Distillation}\label{app:on-policy}

CLPD shares a similar underlying motivation with on-policy distillation in that both aim to mitigate the distribution mismatch between the teacher-generated data and the student’s evolving prediction distribution. In on-policy distillation~\cite{gu2023minillm, agarwal2024policy}, the student actively generates samples from its own policy and queries the teacher for supervision, thereby aligning training data with the student’s current capabilities. In contrast, CLPD achieves a related effect in a fully offline manner by progressively aligning data difficulty with teacher capability. A key aspect of CLPD is the explicit coupling between data difficulty and teacher scheduling: easier samples are paired with weaker teachers at early stages, while harder samples are introduced alongside stronger teachers as training progresses. This coordinated design ensures that the supervision signal remains well-matched to the student’s learning stage, reducing mismatch not only in the data distribution but also in the quality and complexity of teacher guidance. As a result, while on-policy distillation adapts the data distribution dynamically through student sampling, CLPD approximates this adaptation through a structured curriculum over pre-generated data. This perspective highlights CLPD as a stable and efficient alternative that captures key benefits of on-policy distillation without requiring costly online data generation or repeated teacher queries.

This connection also clarifies why CLPD is not directly applicable to on-policy distillation. The primary benefit of CLPD arises from its ability to reduce distribution mismatch in an offline setting through the coupling of data difficulty and teacher scheduling. In contrast, on-policy distillation inherently mitigates this mismatch by constructing training data from the student’s current policy, ensuring that supervision is already well-aligned with the student’s learning state. As a result, the key mechanism that CLPD is designed to address becomes largely redundant in the on-policy setting. Moreover, incorporating teacher scheduling into on-policy distillation is fundamentally different from the offline case. In on-policy settings, switching teachers effectively changes the token-level learning signal (or reward model) during training, rather than merely altering the data distribution. In the absence of the distribution mismatch, it is unclear which teacher provides a more appropriate token-level supervision signal for the student at a given stage. This makes the benefits of teacher scheduling less well-defined and potentially unstable in on-policy distillation. 

We provide additional preliminary results exploring the connection between CLPD and on-policy distillation. We first compare CLPD in the standard off-policy distillation setting with on-policy distillation, and then investigate the effect of applying CLPD within the on-policy framework.

\subsection{CLPD vs. On-policy Distillation}

We first compare CLPD in the standard off-policy distillation setting with on-policy distillation. Both approaches aim to reduce the mismatch between the student’s learning state and the supervision it receives, but they achieve this through fundamentally different mechanisms. On-policy distillation dynamically generates training data from the student’s current policy and queries the teacher for supervision, ensuring that the data distribution is naturally aligned with the student. In contrast, CLPD operates in an offline setting and reduces distribution mismatch through a structured curriculum, achieved by coupling data difficulty with teacher scheduling.

\begin{figure}[t]
    \centering
    \includegraphics[width=0.8\linewidth]{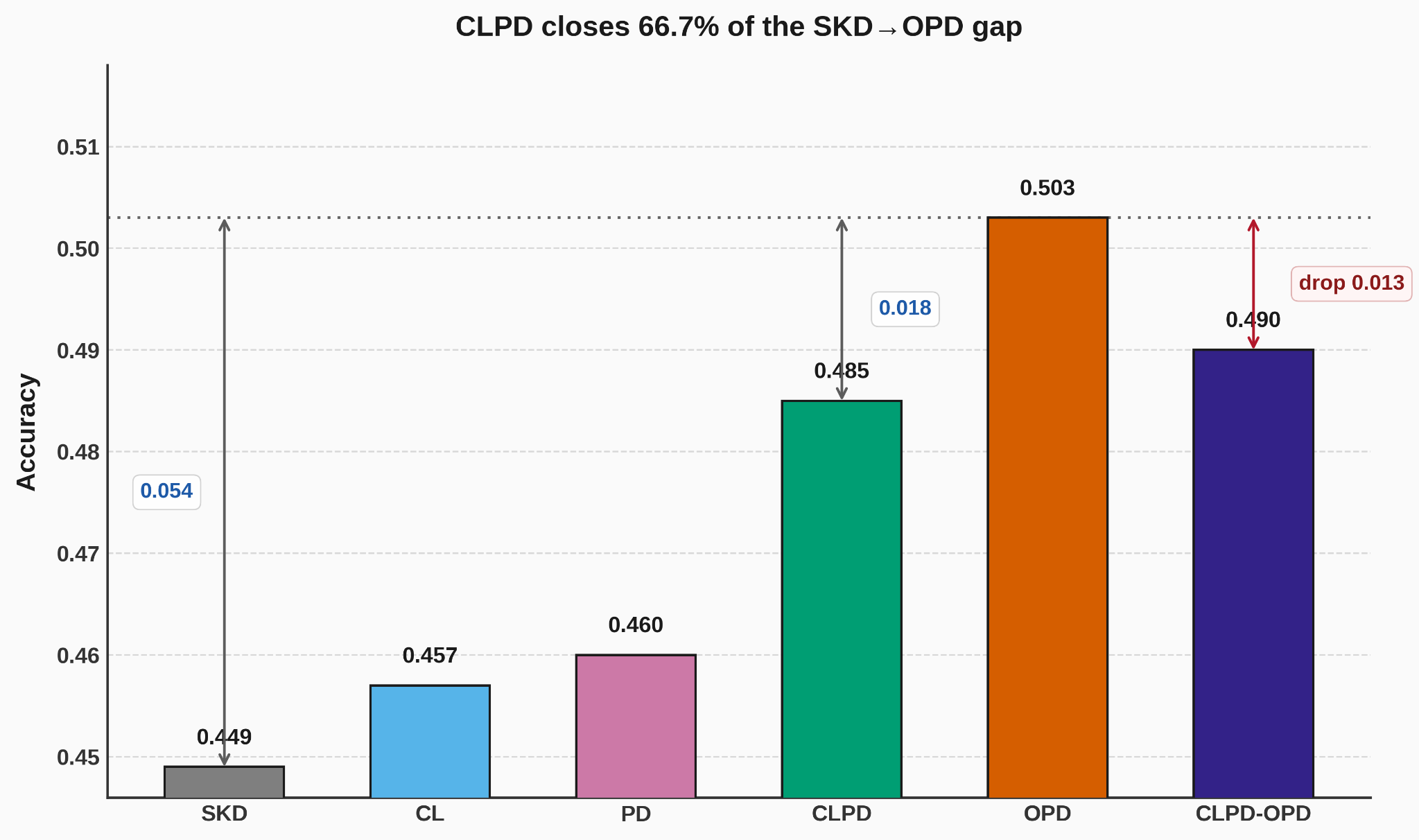}
    \caption{Comparison of CLPD and OPD on GSM8K.}
    \label{fig:OPD}
\end{figure}

Empirically, from Fig.~\ref{fig:OPD}, we observe that CLPD substantially improves over standard off-policy distillation (SKD) and achieves performance that is competitive with on-policy distillation. This suggests that a substantial portion of the gains from on-policy distillation can be attributed to improved alignment between training data and the student’s capability. CLPD provides an efficient alternative by approximating this alignment through a carefully designed curriculum over pre-generated data, without requiring expensive online sampling or repeated teacher queries.

\subsection{Applying CLPD to On-policy Distillation}

We next investigate whether CLPD can further improve performance when applied on top of on-policy distillation. In this setting, the student generates data from its own policy, and the teacher provides supervision on these samples. As a result, the distribution mismatch that CLPD is designed to address is largely mitigated by construction. Nevertheless, we empirically evaluate whether CLPD can still provide additional benefits in this setting.

Fig.~\ref{fig:OPD} also shows that incorporating CLPD into on-policy distillation does not improve performance and can even degrade it. This is consistent with the understanding that the primary benefit of CLPD lies in reducing distribution mismatch in offline settings. Moreover, introducing teacher scheduling in the on-policy setting alters the token-level supervision signal during training, effectively changing the reward model over time. In the absence of a distribution mismatch, it is unclear which teacher provides the most appropriate supervision at each stage. As illustrated in the figure, applying CLPD to the On-Policy Distillation (OPD) setting results in a performance decline. This suggests that for OPD, a single, strong teacher (a token-level reward model) is more effective than a progressive switching strategy.

These findings suggest that CLPD is most effective in offline distillation settings, where supervision mismatch is a primary challenge. In contrast, on-policy distillation already adapts supervision to the student’s evolving policy, leaving limited room for additional gains from curriculum-aware teacher scheduling.

\section{Limitations}\label{app:limitation}

There are several limitations that warrant further investigation. 
(i) CLPD relies on predefined teacher switching points to align data difficulty with teacher capability. While we adopt a simple and effective uniform partition, the optimal switching schedule may depend on the dataset, model pair, and task. Although our ablation studies suggest that CLPD is reasonably robust to partition choices, automatically adapting the switching schedule during training remains an open problem. 
(ii) We primarily evaluate CLPD on mathematical and commonsense reasoning benchmarks, where example difficulty can be naturally characterized and aligned with teacher capability. Extending CLPD to broader domains and more heterogeneous tasks is an important direction for future work. 
(iii) Our experiments mainly focus on standard offline sequence-level distillation. While we discuss the relationship between CLPD and on-policy distillation, the benefits of curriculum-aware teacher scheduling under other sophisticated distillation paradigms remain unclear. In particular, when the training distribution is already adapted to the student policy, the advantages of CLPD may diminish. Exploring these interactions more systematically is an interesting direction for future research.

\section{Detailed Experimental Setup}
\label{app:setup}

This appendix provides detailed information on model configurations (Section~\ref{app:models}), data generation (Section~\ref{app:generated_cot}), training procedures (Section~\ref{app:training}), evaluation protocols (Section~\ref{app:eval}), and baseline methods (Section~\ref{app:baselines}).

\subsection{Base Models}
\label{app:models}

We consider two widely used open-source LLM families—Qwen-2.5 and Llama-3.2—at multiple scales.

\begin{itemize}
    \item \textbf{Qwen-2.5.} A decoder-only Transformer family released by Alibaba. 
    We use \texttt{Qwen2.5-0.5B}, \texttt{Qwen2.5-0.5B-Instruct}, and \texttt{Qwen2.5-3B} as student models, and 
    \texttt{Qwen2.5-3B-Instruct}, \texttt{Qwen2.5-7B-Instruct}, and \texttt{Qwen2.5-14B-Instruct} as teacher models depending on the task.\footnote{\url{https://huggingface.co/Qwen}}
    
    \item \textbf{Llama-3.2.} A decoder-only Transformer family released by Meta. 
    We use \texttt{Llama3.2-1B}, \texttt{Llama3.2-1B-Instruct}, and \texttt{Llama3.2-3B} as student models.\footnote{\url{https://huggingface.co/meta-Llama}}
\end{itemize}

\noindent
\textbf{Teacher selection.}
To control for teacher quality and isolate the effects of curriculum and scheduling, we fix the teacher family to Qwen-2.5 across all main experiments. Empirically, we observe that Llama models underperform Qwen models on the considered reasoning tasks, motivating this choice.

\noindent
\textbf{Student/teacher configurations.}
For GSM8K, the students are \texttt{Qwen2.5-0.5B-Instruct} and \texttt{Llama3.2-1B-Instruct}, and the teachers are \texttt{Qwen2.5-3B-Instruct} and \texttt{Qwen2.5-7B-Instruct}. 
For MATH, the students are \texttt{Qwen2.5-3B} and \texttt{Llama3.2-3B}, and the teachers are \texttt{Qwen2.5-7B-Instruct} and \texttt{Qwen2.5-14B-Instruct}. 
For StrategyQA, the students are \texttt{Qwen2.5-0.5B} and \texttt{Llama3.2-1B}, and the teachers are \texttt{Qwen2.5-3B-Instruct} and \texttt{Qwen2.5-7B-Instruct}. 
For ARC-Challenge, the students are \texttt{Qwen2.5-0.5B} and \texttt{Llama3.2-1B}, and the teachers are \texttt{Qwen2.5-7B-Instruct} and \texttt{Qwen2.5-14B-Instruct}.

\subsection{Generated Chain-of-Thought Data}
\label{app:generated_cot}

All sequence-level distillation experiments use teacher-generated chain-of-thought (CoT) demonstrations as the supervised training targets. To ensure a fair comparison, we first generate and store the teacher outputs offline, and then reuse the same generated data across all methods that use the same teacher. Thus, standard distillation, curriculum learning only, progressive distillation only, and CLPD differ only in the ordering of examples and/or the assignment of teachers, rather than in the underlying teacher generations.

\paragraph{Prompt format.}
For each dataset, we use a fixed prompt template for all teacher models. For mathematical reasoning tasks, the teacher is prompted to solve the problem step by step and provide the final answer in a parsable format. For GSM8K, the final answer is required to be a numeric value. For MATH, the final answer is required to be enclosed in \verb|\boxed{}| when applicable. For commonsense and science reasoning tasks, the teacher is prompted to explain its reasoning step by step and end with a clearly formatted final answer. For StrategyQA, the final answer is restricted to a binary answer, and for ARC-Challenge, the final answer is restricted to one of the candidate option letters. The same prompt template is used for all teachers within each dataset.

\paragraph{Teacher generation.}
Teacher CoT demonstrations are generated using greedy decoding to reduce randomness and improve reproducibility. Specifically, we set the sampling temperature to zero and disable stochastic sampling. The maximum generation length is set according to the task complexity: 512 tokens for GSM8K and StrategyQA, 1024 tokens for MATH, and 256 tokens for ARC-Challenge. These limits match the evaluation-time decoding budget used in Appendix~\ref{app:eval}. Each generated response is stored together with the original example identifier, question, answer label, teacher model name, and the parsed final answer when available. Note that including CoT generally improves performance across many reasoning tasks, including mathematical and commonsense reasoning~\cite{sprague2024cot}.



\paragraph{Data reuse across baselines.}
For single-teacher baselines, all examples are supervised by the same pre-generated CoT outputs from the selected teacher. For progressive distillation and CLPD, each teacher first generates CoT demonstrations for the full training set. The difference between the two methods is how examples are assigned to teachers. Progressive distillation schedules teachers over training while treating the data stream as unordered, whereas CLPD assigns easier curriculum segments to weaker teachers and harder curriculum segments to stronger teachers. Therefore, the comparison isolates the effect of difficulty--capability alignment rather than differences in teacher decoding or prompt design.


\subsection{Training Details}
\label{app:training}

\paragraph{Implementation.}
All experiments are implemented using the TRL framework~\cite{trl2020}. 
We adopt parameter-efficient fine-tuning via LoRA~\cite{hu2022lora}, with rank $r=64$ and scaling factor $\alpha=16$, applied to all attention projection layers. 
Unless otherwise specified, we fine-tune only the LoRA parameters while keeping the base model frozen. 
All training hyperparameters and LoRA configurations are kept fixed across baselines and CLPD variants to ensure fair and controlled comparisons.

\paragraph{Compute.}
All experiments are conducted on \textbf{NVIDIA H100 GPUs}. Training is implemented using PyTorch and the HuggingFace Transformers library with mixed precision (\textbf{bf16}).

\paragraph{Datasets.}
We evaluate CLPD on both mathematical and commonsense reasoning tasks:

\begin{itemize}
    \item \textbf{GSM8K}~\cite{cobbe2021training}: A grade-school math word problem dataset consisting of multi-step arithmetic reasoning questions. Each example requires generating intermediate reasoning steps to arrive at a final numeric answer.

    \item \textbf{MATH}~\cite{hendrycks2024measuring}: A challenging mathematics benchmark covering a wide range of topics (e.g., algebra, geometry, number theory, and combinatorics). Problems are annotated with difficulty levels (Level 1--5) and typically require long-form, structured reasoning.

    \item \textbf{StrategyQA}~\cite{geva2021did}: A binary commonsense reasoning dataset requiring implicit multi-hop reasoning.

    \item \textbf{ARC-Challenge}~\cite{clark2018think}: A multiple-choice science reasoning benchmark focusing on challenging questions.
\end{itemize}

\paragraph{Distillation paradigm.}
For all experiments, we adopt sequence-level knowledge distillation (SeqKD)~\cite{kim2016sequence}, in which the student is trained on teacher-generated chain-of-thought (CoT) outputs via supervised fine-tuning. 
Teacher responses are generated via \textbf{greedy decoding}, and we do not require access to teacher logits.

\paragraph{Optimization.}
We fix the training budget across all methods for fair comparison. Unless otherwise specified:

\begin{itemize}
    \item Optimizer: AdamW
    \item Learning rate: \textbf{2e$^{-4}$} (GSM8K), \textbf{1e$^{-4}$} (MATH, StrategyQA, ARC-Challenge)
    \item Batch size (per device): \textbf{8}
    \item Gradient accumulation steps: \textbf{8}
    \item Number of epochs: \textbf{4} (GSM8K, MATH), \textbf{2} (StrategyQA, ARC-Challenge)
    \item Max sequence length: \textbf{256} {ARC-Challenge},  \textbf{512} (GSM8K, StrategyQA), \textbf{1024} (MATH)
\end{itemize}

\paragraph{Curriculum construction.}
We consider multiple strategies for estimating example difficulty:

\begin{itemize}
    \item \textbf{Expert-based ordering:} Examples are ranked by the number of reasoning steps in expert chain-of-thought (CoT) annotations. This strategy is applicable to datasets such as GSM8K and MATH, where expert CoT is available.

    \item \textbf{Student-loss ordering:} Examples are ranked based on the per-sample loss of a pretrained student model. This approach does not require expert CoT and is applicable to all datasets.
\end{itemize}

\paragraph{Teacher scheduling.}
CLPD schedules teachers from weaker to stronger in alignment with the curriculum (easy-to-hard). 
Unless otherwise specified, we adopt a \textbf{uniform partition} of training steps across teachers as a simple and stable default. 
We additionally report results under non-uniform partitions in Section~\ref{sec:partition} to evaluate the robustness of the proposed method.

\paragraph{Teacher filtering.}
The same viable teacher set is used for all methods. We optionally filter teachers based on task performance using a threshold $\tau$:
\begin{itemize}
    \item $\tau = 90\%$ for ARC
    \item $\tau = 80\%$ for GSM8K
    \item $\tau = 60\%$ for MATH and StrategyQA
\end{itemize}
The threshold $\tau$ is treated as a hyperparameter and can be tuned for each task. In our experiments, we select $\tau$ based on validation performance to optimize CLPD.

In all comparisons, teacher filtering is applied once before training and the resulting teacher set is shared across CLPD and all multi-teacher baselines. Thus, CLPD differs only in how the same teachers are aligned with the curriculum.
\subsection{Evaluation}
\label{app:eval}

\paragraph{Decoding.}
We use a fixed prompt template across all methods. During inference, we apply greedy decoding with a maximum of:
\begin{itemize}
    \item 256 tokens (ARC) 
    \item 512 tokens (GSM8K, StrategyQA)
    \item 1024 tokens (MATH)
\end{itemize}

\paragraph{Metrics.}
We report \textbf{pass@1 accuracy} for all tasks:

\begin{itemize}
    \item \textbf{GSM8K:} Exact match on numeric answers.
    \item \textbf{MATH:} Final boxed answer accuracy using MathVerify.
    \item \textbf{StrategyQA:} Binary accuracy.
    \item \textbf{ARC-Challenge:} Multiple-choice accuracy.
\end{itemize}

\paragraph{Reproducibility.}
All experiments are run with \textbf{3 random seeds}, and we report mean $\pm$ standard deviation in the main results.
We will release code, preprocessing scripts, and evaluation pipelines upon publication.

\paragraph{Low initial performance on commonsense reasoning.}
We observe relatively low initial performance of base models on commonsense benchmarks. This is primarily due to the evaluation protocol: we prompt models to \emph{“Explain your reasoning step by step, then end exactly with the final answer.”} However, pretrained base models often fail to strictly follow this instruction, frequently omitting a clearly formatted final answer at the end of the response, leading to artificially low initial accuracy.

During distillation, the student is trained on teacher-generated chain-of-thought (CoT) demonstrations that consistently adhere to the required format. Consequently, the model quickly learns both to follow the instruction and to produce well-structured reasoning with an explicit final answer, resulting in substantial performance improvements.

\subsection{Baselines}
\label{app:baselines}

We compare CLPD against the following baselines:

\begin{itemize}
    \item \textbf{Standard SeqKD}~\cite{kim2016sequence}: Distillation using a single fixed teacher.
    
    \item \textbf{Curriculum learning only (CL):} Use curriculum ordering but a fixed teacher.
    
    \item \textbf{Progressive distillation only (PD):} Use multiple teachers in increasing capacity, but without curriculum ordering.
    
    \item \textbf{CL+PD (decoupled):} Use both curriculum and multiple teachers, but break the alignment between data difficulty and teacher capacity.
\end{itemize}

\paragraph{Fair comparison.}
All methods share identical student initialization, training budget, optimizer settings, and evaluation protocols.

\section{Qualitative Analysis of Responses from Different Models}

We show the verbatim responses produced by three Qwen models for the same (easy) problem in GSM8K:
\emph{“You can lower the price by 20\% if you buy more than fifteen units of iPhone cases. If you pay \$500 to buy 18 units, what is the original price?”}
All generations are produced with the following prompt.

\begin{prompttemplate}
\textbf{Prompt}

\medskip
Solve the following problem step by step. \\
On the very last line, output exactly: \\
<final\_number>
\end{prompttemplate}

Here are the generations from different models:
\begin{modelbox}[Qwen-2.5-0.5B Instruct]{blue}
Sure! Let's solve this step-by-step.

\textbf{Step 1: Understand the Problem}
\begin{itemize}
    \item You need to find the original price of the iPhone cases.
    \item The new price after a 20\% discount for buying more than 15 units is \$500.
    \item We know that 18 units were bought, which means you paid less than \$500 but more than \$300.
\end{itemize}

\textbf{Step 2: Calculate the Discounted Price per Unit}
\[
\text{Discounted Price per Unit} = \frac{500}{18}
\]

\textbf{Step 3: Determine the Number of Units Bought}

Let $x$ be the number of units you bought. Since you paid less than \$500 but more than \$300:
\[
500 > 18x > 300
\]

\textbf{Step 4: Find the Range for $x$}
\[
x < \frac{500}{18} \approx 27.78, \quad
x > \frac{300}{18} \approx 16.67
\]

\textbf{Step 5: Check the Total Cost}

If $x = 17$, then $17 \times 18 = 306$ (not \$500).  
If $x = 18$, then $18 \times 18 = 324$ (not \$500).  
If $x = 19$, then $19 \times 18 = 342$ (not \$500).  
If $x = 20$, then $20 \times 18 = 360$ (not \$500).  
If $x = 21$, then $21 \times 18 = 378$ (not \$500).  
If $x = 22$, then $22 \times 18 = 396$ (not \$500).  
If $x = 23$, then $23 \times 18 = 314$ (not \$500).  
If $x = 24$, then $24 \times 18 = 324$ (not \$500).  
If $x = 25$, then $25 \times 18 = 360$ (not \$500).  
If $x = 26$, then $26 \times 18 = 378$ (not \$500).

\[
\boxed{500}
\]
\end{modelbox}

\begin{modelbox}[Qwen-2.5-3B Instruct]{green}
To find the original price of the iPhone cases, we follow these steps:

\begin{enumerate}
    \item Determine the discounted price per unit.
    \item Undo the discount to recover the original unit price.
    \item Multiply by the number of units.
\end{enumerate}

The discounted price per unit is:
\[
\frac{500}{18} \approx 27.78
\]

Let $P$ be the original price per unit. Since the discounted price is 80\% of the original:
\[
0.8P = 27.78
\]

Solving for $P$:
\[
P \approx 34.725
\]

The total original price for 18 units is:
\[
18 \times 34.725 = 625
\]

Thus, the original price is:
\[
\boxed{625}
\]
\end{modelbox}

\begin{modelbox}[Qwen-2.5-7B Instruct]{orange}
To solve this problem, let's break it down step by step:

\begin{enumerate}
    \item Identify the given information.
    \item Understand the relationship between the original price and the discounted price.
    \item Set up an equation to represent the relationship.
\end{enumerate}

Since a 20\% discount means the paid price is 80\% of the original price, let $P$ be the original price:
\[
0.8P = 500
\]

Solving for $P$:
\[
P = \frac{500}{0.8} = 625
\]

Thus, the original price is:
\[
\boxed{625}
\]
\end{modelbox}

For this easy instance, all models produce well-formed step-by-step responses; however, closer inspection reveals clear differences in the alignment between the student (0.5B) and each teacher. Although both Qwen-2.5-3B and Qwen-2.5-7B arrive at the correct answer \(\boxed{625}\), the reasoning structure of the 3B teacher is substantially closer to that of the 0.5B student. In particular, both models first compute the discounted unit price as \(500/18\), then explicitly invert the 20\% discount through an equation, and finally scale back to the total original price, even though Qwen-2.5-0.5B ultimately fails to solve this problem correctly. By contrast, the 7B teacher follows a more compressed reasoning path, directly formulating \(0.8P=500\) and solving the problem in a single step. This qualitative comparison supports the observation in Section~\ref{sec:3.3} that, even when multiple teachers produce correct solutions, smaller teachers may induce reasoning trajectories that are more structurally aligned with the student, thereby facilitating more effective distillation on easy instances.



\end{document}